# From Joy to Fear: A Benchmark of Emotion Estimation in Pop Song Lyrics

Shay Dahary
*School of Computer Science, Faculty of Sciences*
Holon Institute of Technology

Avi Edana
*School of Computer Science, Faculty of Sciences*
Holon Institute of Technology

Alexander Apartsin
*School of Computer Science, Faculty of Sciences*
Holon Institute of Technology

Yehudit Aperstein
*Afeka Academic College of Engineering*
Tel Aviv Israel

*Abstract*— The emotional content of song lyrics plays a pivotal role in shaping listener experiences and influencing musical preferences. This paper investigates the task of multi-label emotional attribution of song lyrics by predicting six emotional intensity scores corresponding to six fundamental emotions. A manually labeled dataset is constructed using a mean opinion score (MOS) approach, which aggregates annotations from multiple human raters to ensure reliable ground-truth labels. Leveraging this dataset, we conduct a comprehensive evaluation of several publicly available large language models (LLMs) under zero-shot scenarios. Additionally, we fine-tune a BERT-based model specifically for predicting multi-label emotion scores. Experimental results reveal the relative strengths and limitations of zero-shot and fine-tuned models in capturing the nuanced emotional content of lyrics. Our findings highlight the potential of LLMs for emotion recognition in creative texts, providing insights into model selection strategies for emotion-based music information retrieval applications.

The labeled dataset is available at https://github.com/LLM-HITCS25S/LyricsEmotionAttribution.

## I. Introduction

Music, as a universal form of human expression, evokes a rich spectrum of emotional experiences. While musical elements such as melody and rhythm contribute significantly to emotional perception, the lyrical content often carries the most explicit and interpretable emotional cues. Understanding and quantifying the emotional attribution of song lyrics is a critical task in fields such as music information retrieval, affective computing, and human-computer interaction. Applications of this research span personalized music recommendation systems, emotional playlist generation, mental health interventions, and cultural studies of music trends.

Despite the importance of this task, automatically attributing emotions to lyrics remains a challenging problem. Lyrics often contain abstract language, metaphors, and cultural references that complicate the direct inference of emotions. Traditional sentiment analysis methods, typically designed for binary or ternary sentiment classification, fall short in capturing the multi-dimensional and overlapping emotional content present in lyrical texts. Moreover, the subjective nature of emotional perception necessitates robust labelling methodologies and models that can handle ambiguity and intensity variation across multiple emotional categories.

In this work, we address these challenges by constructing a high-quality, manually labeled dataset for lyric emotion attribution using the Mean Opinion Score (MOS) method. This approach aggregates multiple human annotations to derive reliable emotion intensity scores across six fundamental emotions: joy, sadness, anger, fear, surprise, and disgust. Using this dataset, we systematically evaluate the capabilities of several publicly available large language models (LLMs) under zero-shot settings. Additionally, we develop and fine-tune a BERT-based model tailored explicitly for multi-label emotion score prediction.

Our research provides a comprehensive analysis of model performance across various learning scenarios, highlighting the trade-offs between zero-shot generalization and fine-tuned specialization. By quantifying model effectiveness in the complex task of emotional attribution, this study provides valuable insights into the deployment of language models for affective analysis of creative texts, supporting future advancements in emotionally aware AI systems.

## II. Literature Review

Many modern systems frame emotion categorizations as a multi-label classification problem, recognizing that texts often evoke multiple concurrent emotions. A common approach adopts Ekman's six basic emotions—anger, disgust, fear, happiness (also known as joy), sadness, and surprise—as target labels (Ekman, 1992). Early efforts used lexicon-based or traditional machine learning models, but recent work favours neural methods. For example, Huang et al. (2021) proposed Seq2Emo, a sequence-to-sequence model that decodes correlated emotion labels bidirectionally, outperforming binary relevance baselines on SemEval and GoEmotions benchmarks. Alhuzali and Ananiadou (2021) introduced SpanEmo, casting multi-label emotion tagging as a span-prediction task to capture label dependencies better. Other studies combine deep and handcrafted features; Ahanin et al. (2023) integrated transformer embeddings from BERT, RoBERTa, and XLNet with human-engineered features to improve multi-label accuracy. Classical methods still appear in comparisons; Liu et al. (2023) revisited ML-kNN for short-text emotion tagging, finding that kNN variants are competitive with SVM and Naïve Bayes, but noting that iterative error correction remains challenging. Overall, recent frameworks typically fine-tune transformer encoders on multi-label emotion corpora or employ specialized architectures to model the co-occurrence of emotions, reporting improved F1 and Jaccard scores over traditional approaches.

The analysis of emotions in creative texts, including song lyrics and poetry, has garnered increasing attention. Mihalcea and Strapparava (2008) created an early lyrics dataset labeled with Ekman's emotion categories, demonstrating the complexity of affective analysis in artistic language. Edmonds and Sedoc (2021) expanded this work by introducing the Edmonds Dance dataset, using both Ekman's six emotions and Plutchik's eight-emotion model. They showed that fine-tuning on in-domain lyric data consistently outperforms transferring models trained on social media text. Song and Beck (2022) modelled emotion dynamics in lyrics by treating each song as a time series of sentences. They trained a sentence-level emotion predictor based on Ekman categories and refined predictions using a state-space expectation-maximization algorithm, achieving improved accuracy without requiring fully labeled songs. Addressing the challenge of data scarcity, Sakunkoo and Sakunkoo (2024) proposed cross-domain transfer learning, where a CNN is pre-trained on Reddit comments and then fine-tuned on a small dataset of lyrics. Their results demonstrated that cross-domain knowledge transfer can produce effective emotion models even with limited annotated lyric data.

Large pre-trained transformers, such as BERT and GPT models, are increasingly used for emotion recognition with minimal supervision. Fine-tuning BERT-based models remains a strong baseline, as demonstrated in SpanEmo and other hybrid models (Alhuzali & Ananiadou, 2021). However, recent research explores zero-shot and few-shot learning using large language models. Brown et al. (2020) introduced GPT-3 and demonstrated its ability to perform classification via prompting, a technique later extended to emotion detection. Demszky et al. (2020) created the GoEmotions dataset and fine-tuned BERT for multi-label emotion classification, achieving state-of-the-art performance. Plaza-del Arco et al. (2022) reframed emotion detection as a natural language inference task, using LLMs to determine whether a given text entails a specific emotional hypothesis, achieving strong zero-shot results. Bhaumik and Strzalkowski (2024) proposed EmoGen, leveraging the generative capabilities of T5 to produce open-ended emotion predictions and explanations in a few-shot setting. They found that question-answering style prompting outperformed simple label generation. Boitel et al. (2024) directly compared fine-tuned GPT-3 variants to DeBERTa v3 on standard emotion datasets, concluding that fine-tuned BERT-based models generally outperformed LLM prompting approaches. However, the performance gap narrowed with larger models, such as GPT-4. Current research suggests that while LLMs can achieve competitive results with effective prompting strategies, fine-tuned task-specific models still offer superior performance when labeled data is available.

Building high-quality emotion datasets requires reconciling subjective differences in annotator judgments. The most common strategy for categorical emotion labels is majority voting, as seen in the EmotionLines dialogue corpus, which aggregated annotations from five raters per utterance (Hsu et al., 2018). However, majority voting can obscure individual variations in emotional perception. When emotions are assessed on a continuous scale, researchers often employ the Mean Opinion Score (MOS) method, which involves averaging the ratings of annotators to produce a final intensity score (Chen et al., 2020). Bagdon et al. (2024) argue that direct rating scales can suffer from inconsistency and propose Best-Worst Scaling (BWS) as a more reliable alternative for capturing the intensity of emotions. In speech emotion recognition, the MOS approach remains standard for evaluating perceived emotional strength (Truong & van Leeuwen, 2007). Recent studies also explore probabilistic label models and soft-label aggregation methods to capture better annotator uncertainty and disagreement (Sarumi et al., 2024). Overall, while majority voting and MOS remain dominant strategies, there is growing interest in more nuanced aggregation techniques that preserve the diversity of emotional interpretations across annotators.

### III. METHODOLOGY

#### A. Dataset Construction and Annotation

To establish a reliable ground truth for emotion attribution in song lyrics, we employed a manual annotation process based on the Mean Opinion Score (MOS) methodology. A diverse committee of human annotators, each with prior experience in affective text analysis or music interpretation, participated in the labelling process. Annotators were asked to read each song lyric and independently assign an intensity score ranging from 0 (not present) to 5 (highly present) for each of Ekman's six basic emotions: joy, sadness, anger, fear, surprise, and disgust.

TABLE I
MEAN OPINION SCORE LABELLING BY A COMMITTEE OF FOUR ANNOTATORS: AN EXAMPLE

| Emotion | Annotator 1 | 2 | 3 | 4 | Mean Score (MOS) |
|---|---|---|---|---|---|
| Joy | 2 | 1 | 1 | 1 | 1.25 |
| Sadness | 0 | 1 | 1 | 2 | 1 |
| Anger | 0 | 0 | 0 | 0 | 0 |
| Fear | 0 | 0 | 0 | 0 | 0 |
| Surprise | 1 | 1 | 2 | 2 | 1.5 |
| Disgust | 0 | 0 | 0 | 0 | 0 |

For each lyric and each emotion category, the individual scores were aggregated by computing the arithmetic mean across all annotators, resulting in continuous-valued emotional intensity scores. These MOS values effectively captured the subjective variability among annotators while providing a consistent numerical representation of emotional attribution. This approach also allowed the dataset to reflect the degree of emotional presence rather than forcing binary or categorical labels, which are often inadequate for representing the nuanced emotional content found in song lyrics.

In the resulting dataset, the MOS scores for all emotions are clustered near mid-range values as shown in Figure 1, with the highest mean score corresponding to joy (0.95) and the lowest mean values corresponding to anger (0.40).

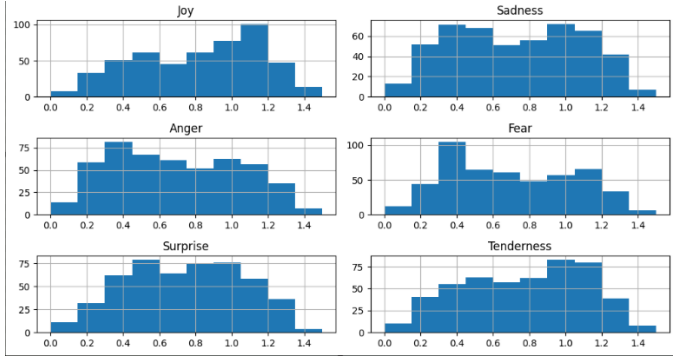

Figure 1: Distribution of scores for each emotion type

There is also no significant correlation between music genre and emotion scores, as shown in Figure 2, with the highest noticeable exception of anger emotion being prominent for rap songs.

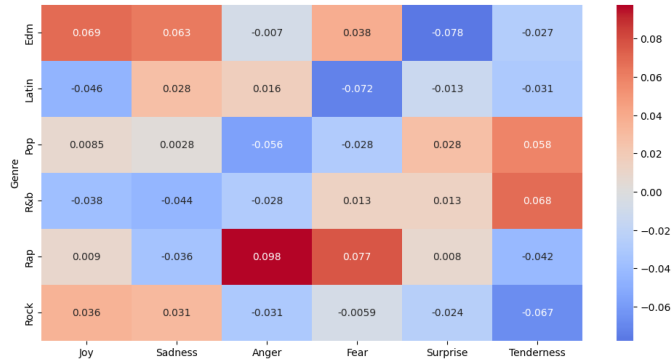

Figure 2: Correlation between music genre and emotion score

### B. Model Evaluation

Using the constructed dataset, we evaluated both fine-tuned and zero-shot models for predicting multi-label emotion scores.

1. *Fine-Tuned BERT-Based Models*: We fine-tuned a pre-trained BERT model for regression-based multi-label prediction, directly predicting the six continuous emotion scores for each lyric. The model architecture consisted of the standard BERT encoder followed by a fully connected regression head with six output nodes, one for each emotion. The loss function used was Mean Squared Error (MSE), which is appropriate for continuous-valued target variables. The model was trained and validated on the annotated dataset to optimize its predictive accuracy across all six emotions.

2. *Zero-Shot Pretrained Models*: In addition to fine-tuned models, we investigated the zero-shot emotion classification capabilities of large language models without requiring further task-specific training using the Grok 3 model.

For zero-shot evaluation, each model was prompted using carefully crafted instructions to predict the emotional intensity scores directly from the raw lyrics. The models were asked to output numeric scores corresponding to the six target emotions. This experimental setup enabled us to evaluate the generalization capabilities of large pre-trained models and their ability to perform fine-grained emotional analysis without requiring specialized fine-tuning.

### IV. RESULTS

To evaluate model performance, we report both aggregated results across all emotions and detailed results for each emotion. Two main evaluation metrics were used: Mean Absolute Error (MAE) and Root Mean Squared Error (RMSE). Lower values indicate better predictive accuracy in estimating the emotional intensity scores.

TABLE II
AGGREGATED RESULTS ACROSS ALL EMOTIONS

| Model | MAE | RMSE |
|---|---|---|
| Finetuned uncase-BERT | 0.33 | 0.1507 |
| Finetuned RoBERTa | 0.33 | **0.1511** |
| Zero-shot Grok LLM | 0.50 | 0.367 |

TABLE III
PER-EMOTION PREDICTION RESULTS

| Emotion | Zero-shot Grok MAE/RMSE | Fine-tuned RoBERTa MAE/RMSE | Fine-Tuned BERT MAE/RMSE |
|---|---|---|---|
| Joy | 0.41/0.55 | **0.17/0.36** | 0.18/0.37 |
| Sadness | 0.42/0.54 | **0.17/0.37** | 0.17/0.37 |
| Anger | 0.45/0.45 | 0.14/0.32 | **0.13/0.31** |
| Fear | 0.35/0.47 | **0.13/0.33** | 0.14/0.32 |
| Surprise | 0.31/0.48 | **0.10/0.26** | 0.10/0.27 |
| Disgust | 0.45/057 | **0.16/0.35** | 0.16/0.35 |

As shown in Table II, both fine-tuned models (uncased BERT and RoBERTa) achieved substantially lower MAE and RMSE values compared to the zero-shot Grok LLM, underscoring the effectiveness of task-specific training on our manually annotated dataset. The performance gap is especially evident in the aggregated metrics, where fine-tuned models reduce error rates by more than 30% relative to the zero-shot approach. A more detailed breakdown in Table III highlights that this improvement is consistent across all six basic emotions. Notably, the fine-tuned models demonstrate strong predictive accuracy for emotions such as *surprise* and *fear*, with MAE values as low as 0.10 and 0.13, respectively. By contrast, the zero-shot model exhibits higher error rates across all categories, particularly for *joy* and *anger*. These results confirm that while zero-shot LLMs provide a baseline capability for lyric emotion attribution, fine-tuned transformer-based models offer far superior performance, making them more suitable for music information retrieval and affective computing applications.

## V. CONCLUSIONS AND FUTURE DIRECTIONS

In this study, we addressed the complex task of emotional attribution in song lyrics by constructing a high-quality, manually labeled dataset using the Mean Opinion Score (MOS) method. This approach enabled us to capture the nuanced and subjective emotional content present in lyrics across Ekman's six basic emotions. Using this dataset, we conducted a comprehensive evaluation of both fine-tuned and zero-shot language models for predicting multi-label emotional intensity.

Our results demonstrate that fine-tuning a BERT-based model significantly outperforms a large zero-shot model such as Grok 3. While zero-shot models offer a practical solution when labeled data is unavailable, their performance remains inferior to models explicitly trained for the task. Among the zero-shot models, DeepSeek-R1 demonstrated stronger generalization capabilities, indicating its enhanced alignment with affective reasoning tasks.

Future work will explore expanding the dataset to include a broader range of lyrical genres and languages, which may improve model generalization and robustness. Additionally, we plan to investigate the effectiveness of advanced prompting strategies, such as chain-of-thought reasoning and structured output prompts, to enhance zero-shot performance. Ultimately, we aim to investigate the application of larger and more specialized language models, as well as multimodal approaches that integrate both lyrics and audio features, to further advance the field of emotion recognition in music.